\documentclass[10pt,twocolumn,letterpaper]{article}

\usepackage{ijcb}
\usepackage{times}
\usepackage{epsfig}
\usepackage{graphicx}
\usepackage{amsmath}
\usepackage{amssymb}
\usepackage{placeins}
\usepackage{url}
\usepackage{adjustbox}
\usepackage[accsupp]{axessibility}

\ijcbfinalcopy 

\ifijcbfinal\fi
\begin{document}



\title{Towards Explaining Demographic Bias through the Eyes of \\ Face Recognition Models}

\author{Biying Fu$^{1,2}$, Naser Damer$^{1,2}$
\\
$^{1}$Fraunhofer Institute for Computer Graphics Research IGD,
Darmstadt, Germany\\
$^{2}$Department of Computer Science, TU Darmstadt,
Darmstadt, Germany\\
{\tt\small biying.fu@igd.fraunhofer.de}
}

\maketitle
\thispagestyle{empty}

\begin{abstract}

Biases inherent in both data and algorithms make the fairness of widespread machine learning (ML)-based decision-making systems less than optimal. 
To improve the trustfulness of such ML decision systems, it is crucial to be aware of the inherent biases in these solutions and to make them more transparent to the public and developers. 
In this work, we aim at providing a set of explainability tool that analyse the difference in the face recognition models' behaviors when processing different demographic groups. We do that by leveraging higher-order statistical information based on activation maps to build explainability tools that link the FR models' behavior differences to certain facial regions.
The experimental results on two datasets and two face recognition models pointed out certain areas of the face where the FR models react differently for certain demographic groups compared to reference groups. The outcome of these analyses interestingly aligns well with the results of studies that analyzed the anthropometric differences and the human judgment differences on the faces of different demographic groups. This is thus the first study that specifically tries to explain the biased behavior of FR models on different demographic groups and link it directly to the spatial facial features. The code is publicly available  here\footnote{\url{https://github.com/fbiying87/Demographic-Bias-Visualization.git}}.

\end{abstract}

\section{Introduction}
\label{sec:introduction}


The performance and accuracy of automated Face Recognition (FR) systems have been boosted lately due to the advances made in deep-learning \cite{DBLP:conf/nips/SunCWT14,DBLP:conf/cvpr/TaigmanYRW14,DBLP:conf/cvpr/SunWT14,DBLP:journals/corr/abs-2109-09416} and large-scaled training face image datasets \cite{guo2016ms, DBLP:conf/fgr/SimBB02, DBLP:conf/cvpr/WolfHM11,DBLP:journals/corr/MillerKS15}. Both algorithmic improvement and large-scaled datasets have contributed to the boom of FR systems being applied in diverse application areas. 
However, research revealed a bias problem in FR Systems. The face recognition vendor test (FRVT) in 2002 \cite{phillips2003face} and later in 2019 \cite{grother2019face} showed that recognition accuracy differs between demographic groups. Among tested FR solutions, some algorithms perform well on Caucasians, while showing less superior performance on other demographics. This biased behavior of the FR solutions causes some problems in diverse applications. While it is less sensitive to make more failure verification on unlocking personal devices, it is more problematic to falsely identify a person as a criminal.
Thus bias funds mistrust in the use of biometric recognition systems both by the scientific communities and the public. Therefore, understanding the biases and making them more transparent could instill trust, fairness, and security into the biometric systems. More importantly, it can help develop new solutions that are specifically designed to be fair.

To contribute toward explaining the demographic bias of FR Models, we propose a set of explainability tools. We show that the average activation mappings of different demographic groups are extremely similar and thus do not reflect the bias. Based on that, we take our explainability tool to a higher derivative of these maps by analyzing the difference (between demographic groups) in the variations in these activation maps.
After motivating our analyses with Fairness analyses on two FR models and two datasets, we demonstrated our explainability pipelines on these models and datasets. 
Our analyses on gender differences and ethnic differences pointed out certain regions of the face where the FR models behave differently in comparison to a reference demographic group.
The results are largely consistent across FR models and datasets.
The results, very interestingly, were consistent with findings in previous studies on facial anthropometric differences and on the human judgment on gender from faces \cite{10.1093/annhyg/meq007,doi:10.1068/p220131}.
This is thus the first attempt to explain the differences in the FR models' behavior on different demographic groups.  
To achieve that, this work provides explaination tools of the FR model behaviour towards a group of samples rather than single samples in the more conventional explainability tools.

\section{Related works} 
\label{sec:related}
The biases inherently built into ML-based systems are a matter of concern, whether data-dependent or human-coded. These biases can be introduced either through data \cite{zhong2019unequal,DBLP:conf/eusipco/FangDKK20}, historical prejudices \cite{oda1997biased}, or other proxies\cite{9412443} which supposedly to be fair representations of the latent bias factors. For example, taking zip codes as a proxy could also include social or ethical bias.

Focusing on fairness/bias in face recognition, there are several works \cite{DBLP:journals/corr/abs-2003-02488, DBLP:journals/tap/PhillipsJNAO11,DBLP:conf/icb/TerhorstKDKK20} pointing out that FR algorithms suffer from the "own-race bias" or the "other-race effect". Drozdowski et al. in \cite{DBLP:journals/corr/abs-2003-02488} pointed out that this effect is visible in most FR algorithms developed in different countries. In \cite{DBLP:journals/tap/PhillipsJNAO11} it is shown, that FR algorithms developed in Asia perform better in recognizing Asian individuals, while solutions developed in Europe perform well on Caucasians. Algorithmic bias in FR tends to over-perform on majority groups and thus making ethnicity a co-variate in the model. But also with balanced ethnicity data, the FR algorithms do not perform equally on all demographic groups, as shown in \cite{klare2012face}. The performance of some groups is still inferior to other groups. This observation gives indications of the inherent and non-quantifiable characteristics of bias. These biases have been shown to extend to other demographic variations, end even non-demographic ones such as personal styling choices \cite{9534882}.

Methods to mitigate these demographic biases are proposed both from the algorithmic viewpoints or data perspective \cite{DBLP:journals/prl/TerhorstKDKK20,wang2021meta,zhong2019unequal,DBLP:conf/iwbf/TerhorstTDKK20}. Wang proposed in \cite{wang2021meta} a Meta-Learning approach to combat the algorithmic bias. The network tries to learn adaptive margins in the latent space for the model to be optimized and perform fairly across people of different skin tones. Later in \cite{wang2020mitigating}, Wang et al. used reinforcement learning to optimize these adaptive margins. From the data perspective, there are works as in \cite{zhong2019unequal, amini2019uncovering}. In \cite{zhong2019unequal} a two-stream approach is used to learn discriminative face representation supervised by mining hard identities on long-tailed data. This iterative way of integrating hard samples from the tail data enables the network to learn through effective batch mining. Amini et al. \cite{ amini2019uncovering} proposed an algorithm for mitigating bias during training by re-sampling the training data according to the automatically learned latent variables within the training stage. The idea is to select rarer data points more often. Other works target balancing the data by adding augmentation of adversarial data per-subject \cite{yucer2020exploring} or adding synthetic data to balance the ethnicity distribution \cite{kortylewski2019analyzing}.

Raising awareness of the bias issue both for the scientific community and the general public is already a start for building fairer and trustworthy solutions \cite{DBLP:journals/corr/abs-2105-14844}. Cross-discipline collaboration of researchers and developers is a mjor requierment to enhance the ML fairness. A major effort in interpreting such phenomena in ML is the explainable artificial intelligence (XAI) program by the Defense Advanced Research Projects Agency \cite{gunning2019darpa}, aiming at promoting innovation in AI in general, not only in privacy-related sectors but also in the fields of medical, finance and autonomous driving. 

\section{Methodology} 
\label{sec:method}
\begin{figure*}
    \centering
    \includegraphics[width=0.7\linewidth]{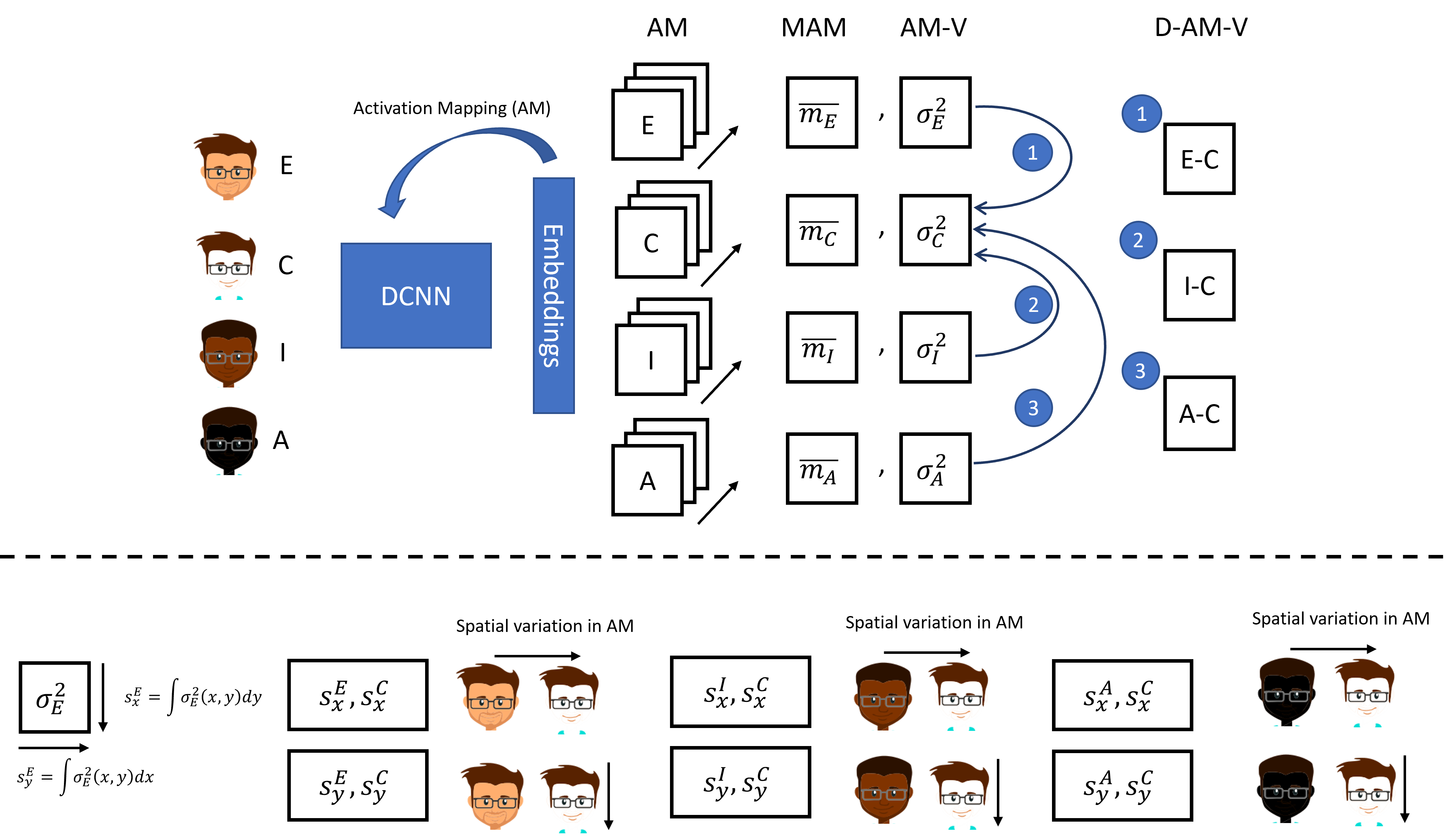}
    \caption{Pipeline illustrates the different processes in our proposed explainability tool. The considered demographic groups are E: East-Asian, C: Caucasian, I: Indian, and A: African. The same processes also apply to gender with males and females. It starts with the activation mapping (AM) from the embedding space for building group statistics using these AM maps. Then the mean activation mapping (MAM) and the activation mapping variation map (AM-V) are determined by building the group mean and group variation of the input AM maps per demographic group. Deviation from the demographic groups is always referred to  Caucasian, as they proved (later in the paper) to lead to the highest FR performance. Differential activation mapping variation (D-AM-V) and the spatial variation map consider the spatial differentiation in activation variation between non-Caucasian and Caucasians. We apply a similar pipeline between the gender groups, female and male.}
    \label{fig:pipeline}
\end{figure*}

Our explainability toolset is built upon the activation mappings (AM) of the FR solutions. These AMs are used to create heat maps for the input image, highlighting the important regions in terms of the network's output. However, from one side, these heat maps deviate largely between samples (due to variations in pose, expression, illumination, etc.), which makes making general conclusions on the effect of different demographic groups virtually pointless. On the other side, they are almost identical if averaged on large groups of samples, even if each sample represents a different demographic group, which limits their explainability utilization as will be shown later in this work. To avoid this, our explainability tools go beyond the base activation mappings into a higher derivative where one can notice statistical differences between groups of face samples. A similar concept has been lately used to derive reasoning for differences between face images of different qualities \cite{DBLP:conf/wacv/FuD22}. These explainability tools are presented in this section and demonstrated in Figure \ref{fig:pipeline}.


\subsection{Activation CAM method}

As a backbone of our explainability tool, we require a method to represent the special activation properties induced by a single sample in FR models. The AM visualization scheme used in this work is the Score-CAM proposed by Wang et al. in \cite{wang2020score}. This method is designed to efficiently display visual explanations for CNNs. It re-weighted the final activation based on emphasizing the most relevant regions within each feature map according to the network's decision. These activation CAM methods surpass the inherent limitations in the gradient-based CAMs \cite{selvaraju2017grad} and provide a more effective and faster way to calculate the salient map \cite{ramaswamy2020ablation}.

\subsection{Our proposed explainability tools}

Figure\,\ref{fig:pipeline} illustrates the different processes that comprised our explainability tools below. 
For each input face image, the Score-CAM generates an output AM with respect to the two FR solutions. 
Each pixel value is denoted as $a_{i,j}$ with $\{i=1:112, j=1:112\}$. This saliency map is the up-sampled and re-weighted activation of the output feature layers according to the penultimate layer of the FR model. This penultimate layer is placed before the FC layer of the FR model. This layer is originally used to generate identity descriptors. For symmetry reasons, we also include the AM of the horizontally flipped image in our calculations. The exact FR models used in this work will be presented in Section \ref{sec:face_recognition}.

We introduce the mean activation mappings (MAM) for each of the demographic groups. We denote them as $\text{MAM}_{dg/g}$ with $dg=\{E,C,I,A\}$ for demographic groups which include the Asian, Caucasian, Indian, and African and $g=\{m,f\}$ for gender and includes males and females. Each element in the MAM is denoted as $\overline{a_{i,j}}$ and it is derived from Eq. (\ref{eq:mam}) using the activation value of each single AM $a_{i,j}$: 

\begin{equation}
    \overline{a_{i,j}}=\frac{1}{N} \sum_{k=1}^{N} a_{i,j}^k ,
    \label{eq:mam}
\end{equation}
where $N$ is the number of images within each demographic per database. The MAM will be shown later (in Section \ref{sec:results}) to have no comprehensive pattern to study the FR behavior differences between demographic groups.

To take our explainability tool to a space where we can expect higher order differences between demographic groups, we measure the variability in the AM, resulting in the activation mapping variation map (AM-V) as higher order analysis, where each pixel is denoted as $\sigma^2_{i,j}$ and is determined in Eq. (\ref{eq:am-d}): 
\begin{equation}
    \sigma^2_{i,j}=\sqrt{\frac{1}{N} \sum_{k=1}^{N} (a_{i,j}^k-\overline{a_{i,j}})^2} ,
    \label{eq:am-d}
\end{equation},
where $N$ is the number of samples in each demographic per-database and $a_{i,j}$ is the element of the individual AM. AM-V thus aims at spatially showing the degree of variation in the activation of a group of samples.

MAM and AM-V look into the spatial areas where either a high activation or a relatively large variation in the activation of the FR occur, respectively. However, as we will see later, the differences between the MAM of images per demographic group do not reveal a lot of explainability information. Therefore, to uncover the spatially related difference between these demographic groups, we need to analyze the differences between the variations of activation in higher derivatives. We introduce the term Differential activation mapping variation (D-AM-V) as in Eq. (\ref{eq:D-AM-V}) 
\begin{equation}
    \text{D-AM-V} = |\text{AM-V}_{dg,1} - \text{AM-V}_{dg,2}| ,
    \label{eq:D-AM-V}
\end{equation}
where the term is calculated between two different demographic groups. Due to the symmetry constraint, this mapping is further mirrored and averaged to enhance the left-and-right symmetry of a face image. The D-AM-V thus graphically highlights the facial areas where two different demographic groups have large differences in their activation variations.

To further enable easier conclusions from the AM-V maps, we further integrate the AM-V map along both x- and y-direction to illustrate the spatial variations along the horizontal and vertical face axes by Eq.\,(4) and (\ref{eq:spatial_var}), namely the spatial-variation-x ($ s_{x}^{dg}$) and spatial-variation-y ($ s_{y}^{dg}$). Using this measure of two demographic groups further provides locations with higher activation variation differences between them. 

\begin{alignat}{1}
    s_{x}^{dg} &= \int_{1}^{112}\text{AM-V}(x,y)\,dy \\
    s_{y}^{dg} &= \int_{1}^{112} \text{AM-V}(x,y)\,dx 
    \label{eq:spatial_var}
\end{alignat}   

The details of the considered pairs of demographic groups will be discussed in more detail in the next section.

\begin{figure*}
\centering
    \begin{tabular}{ccc}
     (a) FDR on RFW over 4 ethnicity & (b)  FDR on BFW over 4 ethnicity  & (c) FDR on BFW gender\\
\includegraphics[width=0.25\textwidth]{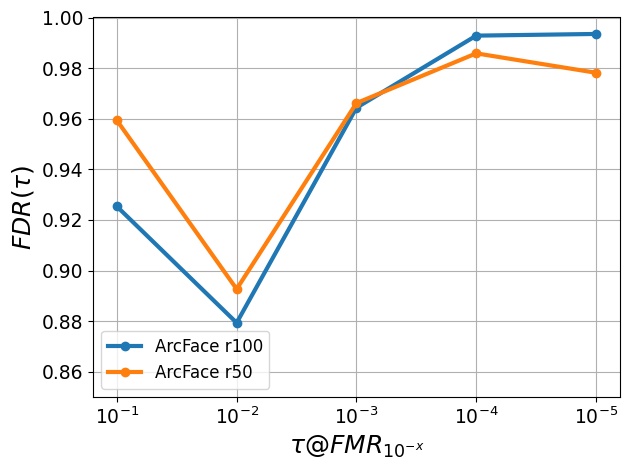} &
    \includegraphics[width=0.25\textwidth]{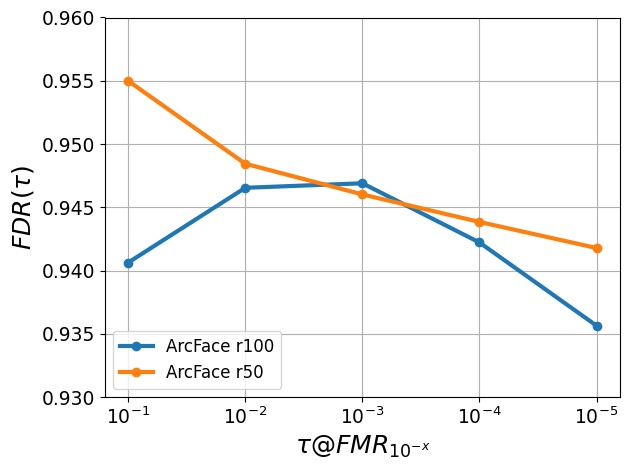} &
    \includegraphics[width=0.25\textwidth]{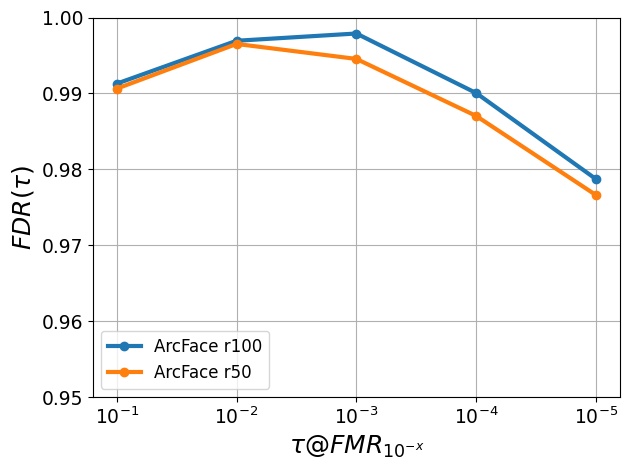}\\ 
\end{tabular}
    \caption{Fairness Discrepancy Rate (FDR) of two FR models over multiple decision thresholds $\tau$ to compare the considered FR models over ethnicity and gender groups.}
    \label{fig:fdr_database}
\end{figure*}

\begin{table*}
\centering
\begin{adjustbox}{width=.7\linewidth}
\begin{tabular}{l|llllll|llllll}
\cline{2-13}
            & \multicolumn{6}{c|}{RFW}                                                                    & \multicolumn{6}{c}{BFW}                                                               \\ \cline{2-13} 
            & \multicolumn{3}{c|}{ArcFace r100}                                                       & \multicolumn{3}{c|}{ArcFace r50}                                   & \multicolumn{3}{c|}{ArcFace r100}                                                       & \multicolumn{3}{c}{ArcFace r50}                                    \\ \hline
         $\tau@FMR_{10^x}$   & \multicolumn{1}{l|}{$10^{-2}$}      & \multicolumn{1}{l|}{$10^{-3}$}      & \multicolumn{1}{l|}{$10^{-4}$}      & \multicolumn{1}{l|}{$10^{-2}$}      & \multicolumn{1}{l|}{$10^{-3}$}      & $10^{-4}$      & \multicolumn{1}{l|}{$10^{-2}$}      & \multicolumn{1}{l|}{$10^{-3}$}      & \multicolumn{1}{l|}{$10^{-4}$}      & \multicolumn{1}{l|}{$10^{-2}$}      & \multicolumn{1}{l|}{$10^{-3}$}      & $10^{-4}$      \\ \hline
         \hline
Demographic & \multicolumn{3}{c|}{FMR}                                                                & \multicolumn{3}{c|}{FMR}                                           & \multicolumn{3}{c|}{FMR}                                                                & \multicolumn{3}{c}{FMR}                                            \\ \hline
Caucasian   & \multicolumn{1}{l|}{1.5E-1} & \multicolumn{1}{l|}{1.0E-2} & \multicolumn{1}{l|}{1.6E-3} & \multicolumn{1}{l|}{1.9E-1} & \multicolumn{1}{l|}{1.5E-2} & 0.7E-3 & \multicolumn{1}{l|}{0.7E-2} & \multicolumn{1}{l|}{0.4E-3} & \multicolumn{1}{l|}{0.1E-4} & \multicolumn{1}{l|}{0.9E-2} & \multicolumn{1}{l|}{0.6E-3} & 0.4E-4 \\ \hline
Asian       & \multicolumn{1}{l|}{3.8E-1} & \multicolumn{1}{l|}{8.0E-2} & \multicolumn{1}{l|}{12E-3}  & \multicolumn{1}{l|}{4.1E-1} & \multicolumn{1}{l|}{7.4E-2} & 9E-3   & \multicolumn{1}{l|}{2.6E-2} & \multicolumn{1}{l|}{3.6E-3} & \multicolumn{1}{l|}{3.9E-4} & \multicolumn{1}{l|}{1.9E-2} & \multicolumn{1}{l|}{2.6E-3} & 3.1E-4 \\ \hline
African     & \multicolumn{1}{l|}{4.2E-1} & \multicolumn{1}{l|}{9.0E-2} & \multicolumn{1}{l|}{17E-3}  & \multicolumn{1}{l|}{5.1E-1} & \multicolumn{1}{l|}{10E-2}  & 13E-3  & \multicolumn{1}{l|}{1.7E-2} & \multicolumn{1}{l|}{1.8E-3} & \multicolumn{1}{l|}{2.2E-4} & \multicolumn{1}{l|}{1.9E-2} & \multicolumn{1}{l|}{2.2E-3} & 2.1E-4 \\ \hline
Indian      & \multicolumn{1}{l|}{3.9E-1} & \multicolumn{1}{l|}{8.1E-2} & \multicolumn{1}{l|}{11E-3}  & \multicolumn{1}{l|}{4.6E-1} & \multicolumn{1}{l|}{9.0E-2} & 8E-3   & \multicolumn{1}{l|}{2.2E-2} & \multicolumn{1}{l|}{2.1E-3} & \multicolumn{1}{l|}{1.8E-4} & \multicolumn{1}{l|}{2.3E-2} & \multicolumn{1}{l|}{2.4E-3} & 2.4E-4 \\ \hline
\hline
            & \multicolumn{3}{c|}{FNMR}                                                               & \multicolumn{3}{c|}{FNMR}                                          & \multicolumn{3}{c|}{FNMR}                                                               & \multicolumn{3}{c}{FNMR}                                           \\ \hline
Caucasian   & \multicolumn{1}{l|}{0.7E-3} & \multicolumn{1}{l|}{0.005}  & \multicolumn{1}{l|}{0.008}  & \multicolumn{1}{l|}{0.003}  & \multicolumn{1}{l|}{0.013}  & 0.043  & \multicolumn{1}{l|}{0.040}  & \multicolumn{1}{l|}{0.057}  & \multicolumn{1}{l|}{0.090}  & \multicolumn{1}{l|}{0.044}  & \multicolumn{1}{l|}{0.073}  & 0.112  \\ \hline
Asian       & \multicolumn{1}{l|}{1.3E-3} & \multicolumn{1}{l|}{0.004}  & \multicolumn{1}{l|}{0.012}  & \multicolumn{1}{l|}{0.004}  & \multicolumn{1}{l|}{0.022}  & 0.063  & \multicolumn{1}{l|}{0.127}  & \multicolumn{1}{l|}{0.160}  & \multicolumn{1}{l|}{0.205}  & \multicolumn{1}{l|}{0.137}  & \multicolumn{1}{l|}{0.178}  & 0.224  \\ \hline
African     & \multicolumn{1}{l|}{0.7E-3} & \multicolumn{1}{l|}{0.002}  & \multicolumn{1}{l|}{0.005}  & \multicolumn{1}{l|}{0.002}  & \multicolumn{1}{l|}{0.011}  & 0.037  & \multicolumn{1}{l|}{0.085}  & \multicolumn{1}{l|}{0.109}  & \multicolumn{1}{l|}{0.136}  & \multicolumn{1}{l|}{0.090}  & \multicolumn{1}{l|}{0.121}  & 0.155  \\ \hline
Indian      & \multicolumn{1}{l|}{0.7E-3} & \multicolumn{1}{l|}{0.003}  & \multicolumn{1}{l|}{0.006}  & \multicolumn{1}{l|}{0.003}  & \multicolumn{1}{l|}{0.014}  & 0.044  & \multicolumn{1}{l|}{0.082}  & \multicolumn{1}{l|}{0.105}  & \multicolumn{1}{l|}{0.131}  & \multicolumn{1}{l|}{0.087}  & \multicolumn{1}{l|}{0.119}  & 0.151  \\ \hline
\hline
FDR         & \multicolumn{1}{l|}{0.879}  & \multicolumn{1}{l|}{0.964}  & \multicolumn{1}{l|}{0.992}  & \multicolumn{1}{l|}{0.892}  & \multicolumn{1}{l|}{0.966}  & 0.985  & \multicolumn{1}{l|}{0.946}  & \multicolumn{1}{l|}{0.946}  & \multicolumn{1}{l|}{0.942}  & \multicolumn{1}{l|}{0.948}  & \multicolumn{1}{l|}{0.946}  & 0.943  \\ \hline
FDR AUC     & \multicolumn{3}{l|}{0.949}                                                              & \multicolumn{3}{l|}{0.953}                                         & \multicolumn{3}{l|}{0.943}                                                              & \multicolumn{3}{l}{0.947}                                         \\ \hline
\end{tabular}
\end{adjustbox}
\caption{FNMR($\tau$), FMR($\tau$), and FDR($\tau$) are given per demographic group, where the operational points are defined as $\tau$ at $FMR_x$. It is to note that $\tau$ is set using the entire test dataset as a global threshold.}
\label{tab:performance_ethnicity}
\end{table*}

\section{Experimental Setup} 
\label{sec:experiment}

This section provides an overview of our experimental setup in terms of the ethnicity and gender-balanced face datasets, fairness evaluation metrics, considered FR models, and the investigated demographic differences.

\subsection{Database}
\label{sec:datasets}
We performed our experiment on two publicly available face datasets especially designed for validating the demographic bias in FR algorithms. As opposed to other large-scaled face image datasets with heavily unbalanced and long-tailed distributions, these two datasets have a balanced number of subjects in each of the four ethnicity groups included.

Robinson et al. proposed the Balanced Faces in the Wild (BFW) in \cite{robinson2020face}. The data consists of four different ethnic groups (Asian, Black, Indian, and White). Each ethnicity is further split into two subgroups of balanced males and females. Each subgroup has 25 faces of 100 subjects and aggregated to 20K faces in total. This dataset is used both for the investigation of ethnicity bias and gender bias, where we combined all female and male subjects across all ethnicity groups. Five folds cross-validation is used in the BFW dataset, with in total more than 920K pairs of 240K genuine and 680K imposter comparisons. Based on these comparison pairs, we separate them further into Caucasian-Caucasian, Asian-Asian, African-African, and Indian-Indian pairs, as well as the female-female and male-male pairs. 

The Racial Faces in-the-wild (RFW) in \cite{wang2019racial} also consists of four testing ethnicity groups, namely Caucasian, Asian, Indian, and African. Each subset contains around 10K images of 3K individuals. In RFW, the images are carefully balanced and cleaned. As no gender labels are provided for this dataset, we only use this dataset for the investigation of FR models on ethnicity differences. RFW dataset composes of 6000 pairs of equal genuine (3000) and imposter (3000) pairs for each ethnicity (Caucasian-Caucasian, Asian-Asian, African-African, and Indian-Indian pairs), which makes in a total of 24K pairs of genuine and imposter comparisons. 

\subsection{Face recognition models}
\label{sec:face_recognition}

Our experiments are performed on two FR models. Both FR models are trained with ArcFace loss \cite{DBLP:conf/cvpr/DengGXZ19} and publicly released by their creators \footnote{https://github.com/deepinsight/insightface}. Both backbones of the FR models are based on the ResNet architecture \cite{DBLP:conf/cvpr/HeZRS16} of different scales. The larger backbone is the ResNet-100, which has deeper middle layers for the feature extraction compared to the smaller backbone with ResNet-50. This selection is motivated by the effect of the model scale on ML bias pointed out in \cite{DBLP:journals/corr/abs-2010-03058}. We chose the ArcFace r100 model, as the solution shows high performance on face identification accuracy, across substantial changes in viewpoint, illumination, expression, and quality \cite{DBLP:conf/cvpr/DengGXZ19, mallick2022influence}. We base our investigations on FR models of different scales to further show the behavioral patterns of demographic fairness across models of different scales. 

To match the input expected by the FR models, each face image is first cropped and aligned (similarity transfer) using MTCNN \cite{DBLP:journals/spl/ZhangZLQ16, DBLP:conf/cvpr/DengGXZ19} into standardized face images of $112\times112$ pixels. For both ArcFace models, the output layer \textit{net.layer4} is used for activation mapping. To mitigate the non-symmetry issue, we include both the horizontally flipped version of the input image and its original version for the activation mapping. 

\subsection{Fairness evaluation metrics}

To measure the fairness of the FR models, as a motivation for our explainability efforts, we adopt the  Fairness discrepancy ratio (FDR) proposed in \cite{de2021fairness}. The FDR takes both verification errors, namely the false match rate (FMR) and the false non-match rate (FNMR) into consideration for a given decision threshold. A biometric verification system is said to be fair only if, at a given decision threshold, statistical equality can be achieved for all pairs of demographic groups in terms of both FMR and FNMR. A higher FDR value indicates a fairer behavior between two demographic groups. The equation of FDR is given by Eq. 
\begin{equation}
    FDR(\tau) = 1 - (\alpha A(\tau) + (1-\alpha)B(\tau)),
    \label{eq:fdr}
\end{equation}
where the term $A(\tau)$ and $B(\tau)$ are the two premises in \cite{de2021fairness} considering both the FMR and FNMR measures. The equations are $A(\tau) = max(|\text{FMR}^{dg_i}(\tau) - \text{FMR}^{dg_j}(\tau)|)\leq\epsilon$ and $B(\tau) = max(|\text{FNMR}^{dg_i}(\tau) - \text{FNMR}^{dg_j}(\tau)|)\leq\epsilon$, where $dg_i,dg_j$ are each from one demographic group. $\alpha$ is set to $0.5$ in our experiments, giving both error types an equal contribution to FDR. $\epsilon$ is a relaxation constraint that puts a limit of when to consider a system "fair" \cite{de2021fairness}, which we set in our analyses to zero.

To better compare the two FR models, we plot the FDR as a function of an operational threshold $\tau$. The global threshold $\tau$ is determined on the entire test dataset from all demographics following \cite{de2021fairness}. For the FDR curve, we vary $\tau$, the global threshold that results in an FMR of $10^{-1}$ to $10^{-5}$ in 5 steps, the $\tau$ will be noted by the FMR threshold is calculated. We also provide the area under the curve of FDR (within the same range) as another measure to assess the fairness of a certain FR model. 

\subsection{Investigation}

The experiments are designed to address the two main demographic variations of our study (1) ethnic differences, and (2) gender differences.

Before introducing the findings of the research scope, we first motivate the need by looking at the demographic fairness issue in both considered FR models in terms of verification performance and FDR metric. Then, we applied our proposed explainability tools to the BFW and RFW datasets using both FR models of different scales to back-propagate the network's decision via activation mapping on the input data. Both FR models have the same DCNN-based backbone ResNet-100 and RestNet-50. We build our analysis on these sets of activation mappings. 
As we only have gender labels for the BFW dataset, the experiments on gender bias addressing the second aspect are only conducted on the BFW dataset. Similarly, we applied our chain of tools to the gender-balanced dataset to draw implications on the network's commonality on the gender aspect. 

As both FR solutions are trained on face datasets with Caucasian males as the majority class and as will be demonstrated, perform the best on Caucasians and males, the experiments conducted in this study compare the demographic groups against the Caucasian group when considering demographics as a reference, and the female group against the male group as a reference when considering gender.

\begin{figure*}
    \centering
    \includegraphics[width=.8\textwidth]{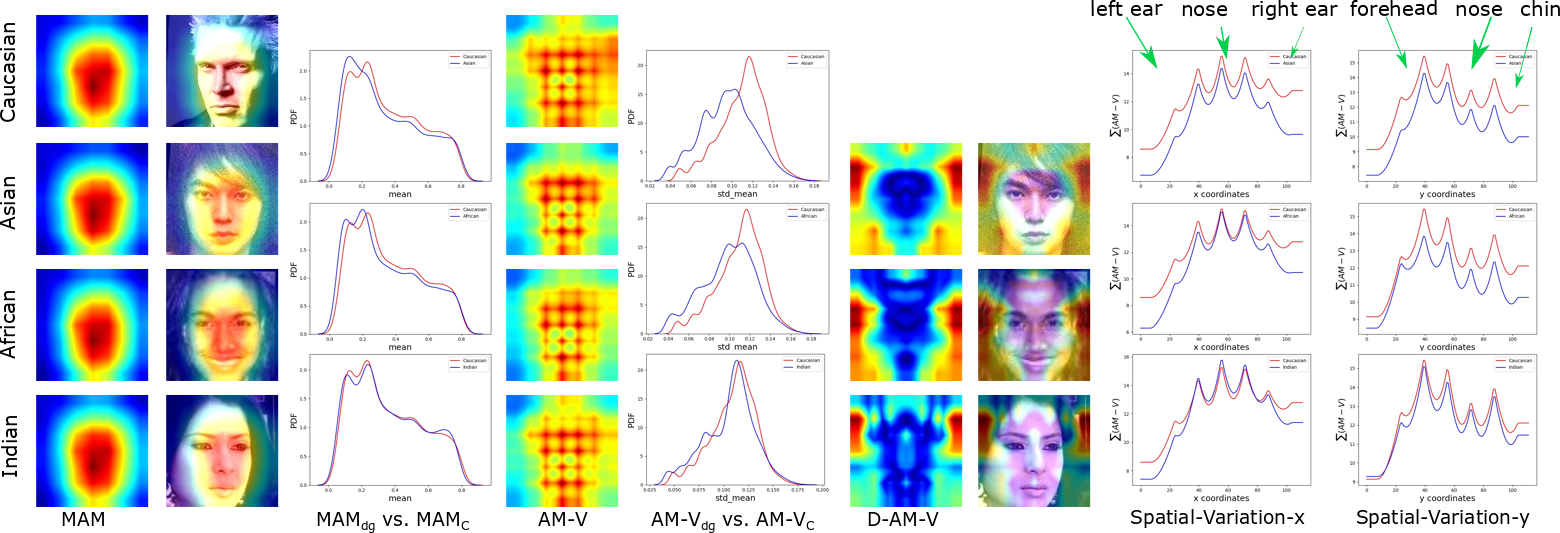}
    \caption{The Explainability tool is applied to all four ethnic groups in the BFW dataset, using ArcFace r100 as the FR model. The 1st column shows the MAM of each ethnic group, while the 2nd column shows the MAM map superimposed on a sample image of each group. The 3rd column shows the distribution of the values of the MAM map of two groups. The 4th column is the AM-V. The 5th column visualizes the distribution of AM-V values in two demographic groups. The D-AM-V maps in column 6 and overlaid with sample images in column 7, showing the main areas that trigger different behavior in the FR models. The last two columns, 8 and 9, summarize the AM-V by summing its values on the horizontal and vertical face axes for two different demographic groups. Higher values indicate higher local variation in activation for this ethnic group.
    }
    \label{fig:ArcFace_BFW_scoreCAM}
\end{figure*}

\begin{figure*}
    \centering
    \includegraphics[width=.8\textwidth]{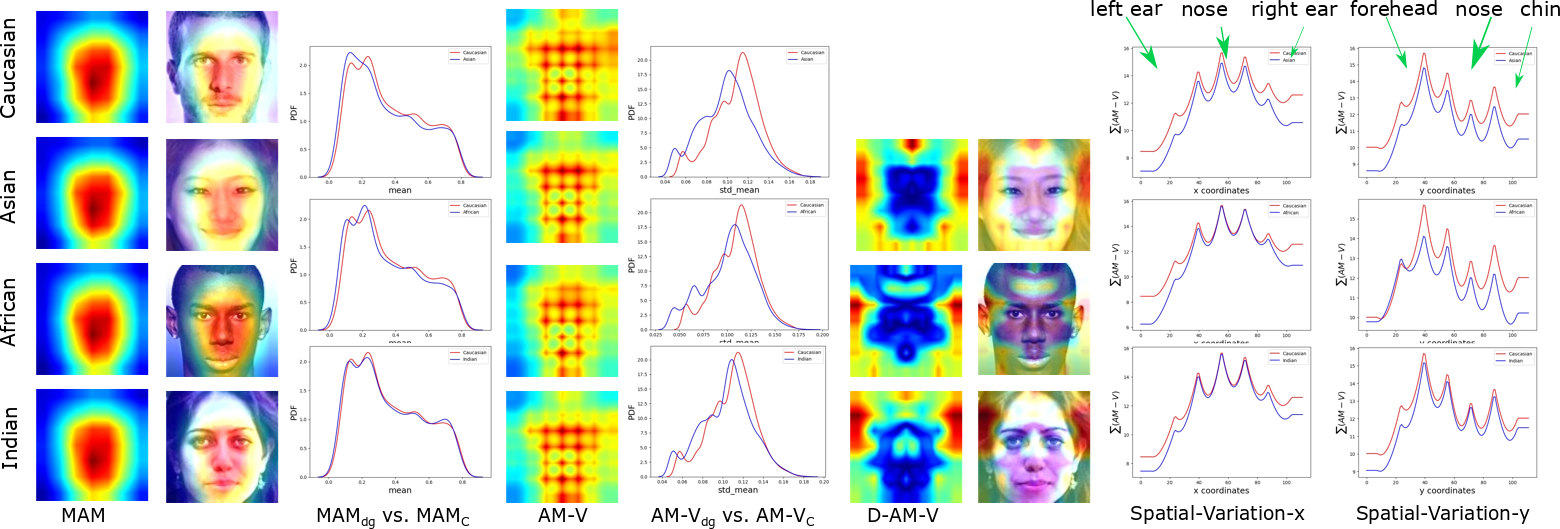}
    \caption{The Explainability tool is applied to all four ethnic groups in the RFW dataset, using ArcFace r100 as the FR model. The different columns follow the explanation in Figure \ref{fig:ArcFace_BFW_scoreCAM} caption.}
    \label{fig:ArcFace_RFW_scoreCAM}
\end{figure*}

\begin{figure*}
    \centering
    \includegraphics[width=.8\textwidth]{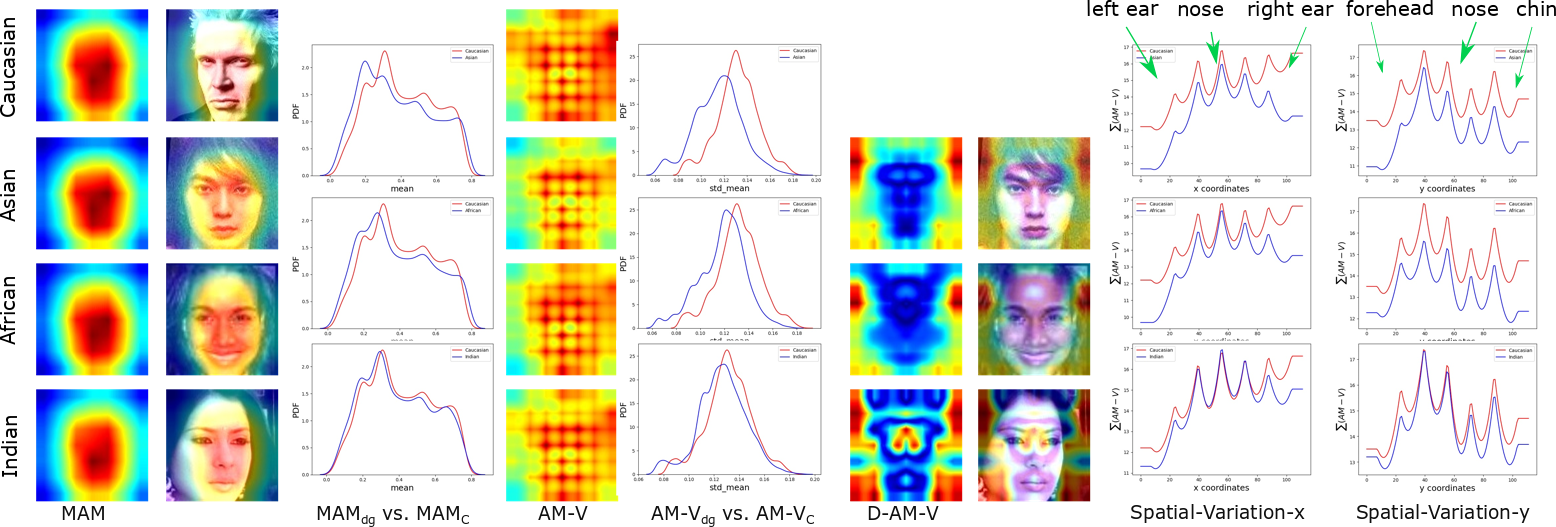}
    \caption{The Explainability tool is applied to all four ethnic groups in the BFW dataset, using ArcFace r50 as the FR model. The different columns follow the explanation in Figure \ref{fig:ArcFace_BFW_scoreCAM} caption.}
    \label{fig:ResNet50_BFW_scoreCAM}
\end{figure*}

\begin{figure*}
    \centering
    \includegraphics[width=.8\textwidth]{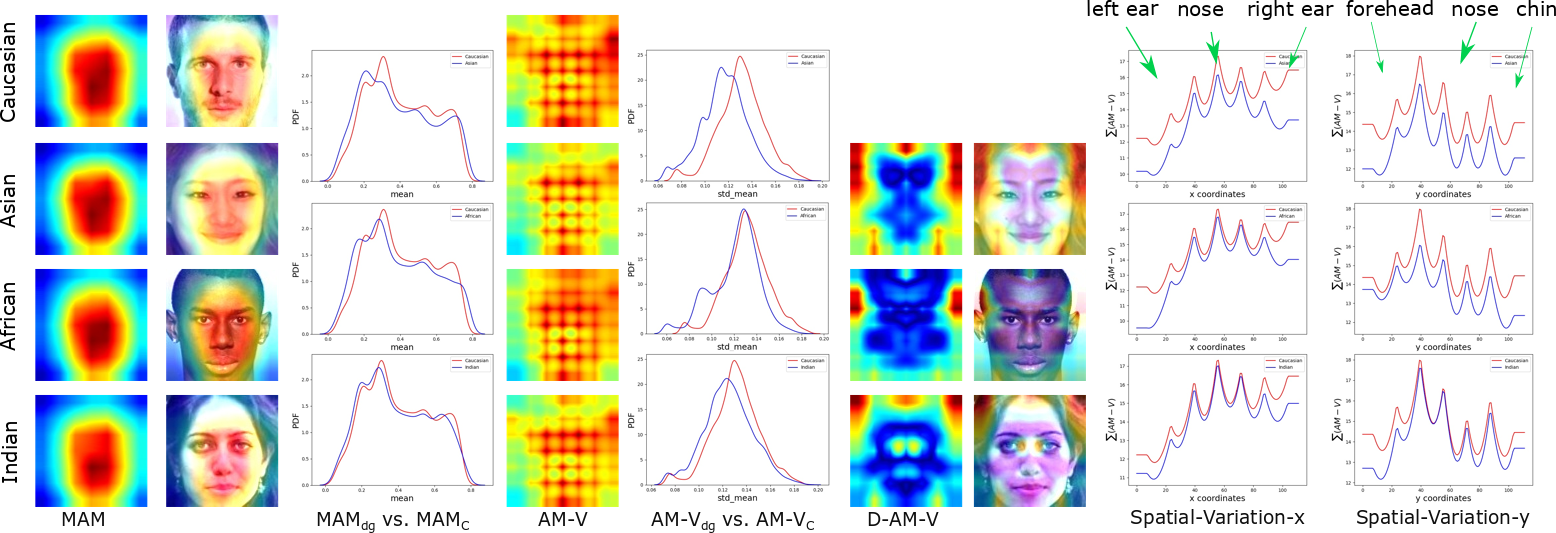}
    \caption{The Explainability tool is applied to all four ethnic groups in the RFW dataset, using ArcFace r50 as the FR model. The different columns follow the explanation in Figure \ref{fig:ArcFace_BFW_scoreCAM} caption.}
    \label{fig:ResNet50_RFW_scoreCAM}
\end{figure*}

\begin{figure*}
    \centering
    \includegraphics[width=.8\textwidth]{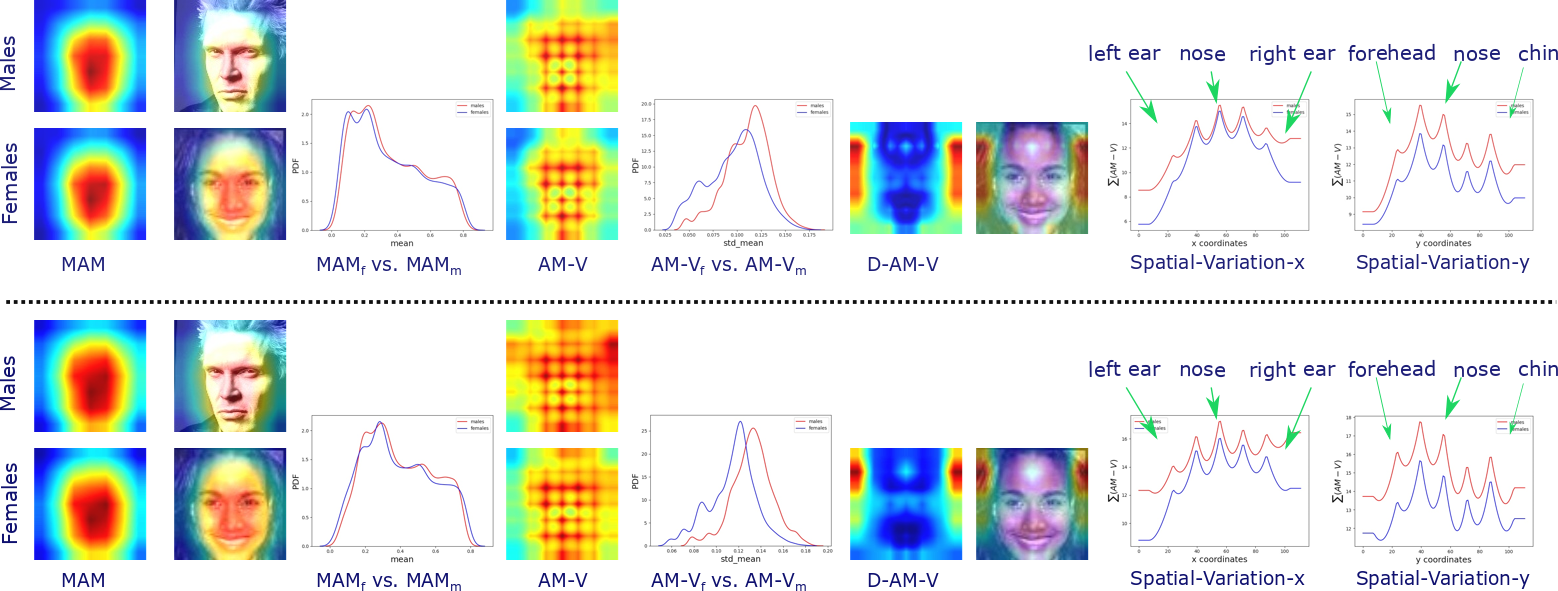}
    \caption{The Explainability tool is applied to both gender demographic groups in the BFW dataset, using ArcFace r100 (upper) and ArcFace r50 (below) as the FR model. The different columns follow the explanation in Figure \ref{fig:ArcFace_BFW_scoreCAM} caption.}
    \label{fig:gender_BFW_ScoreCAM}
\end{figure*}

\section{Results}
\label{sec:results}

In this section, we discuss the results of the two explainability aspects (ethnic and gender differences) using our presented tools with respect to both considered FR models.

\subsection{Demographic fairness}

Demographic fairness requires the automatic FR algorithms to perform equally on all different demographic groups for any $\tau$. However, recent studies \cite{grother2019face,9534882} show that depending on the underlying FR model, the system does not perform equally well for all ethnic groups. 

Here, to build a basis for our explainability analyses, we analyse the fairness of the considered FR models. We first discuss the fairness in demographics in terms of verification performance comparison for our datasets. Table \ref{tab:performance_ethnicity} shows that both FR models produce different performances for different ethnicity groups. 
For the BFW benchmark, both FR models performed the best (in terms of FMR and FNMR) on faces from the Caucasian demographic group.
Other ethnicity groups in BFW do perform worse than the Caucasian group by scoring higher FNMR and FMR values at most global operational thresholds. 
Similar conclusion can be made from the RFW benchmarks, especially from the FMR values, with the FNMR is slightly less consistent due to the lower number of the genuine pairs (lower statistical significance) in the RFW benchmark in comparison to BFW (See Section \ref{sec:datasets}).
In general, ArcFace r100 outperforms the smaller model ArcFace r50, as expected, in most experimental settings on both the BFW and RFW benchmarks.

Fairness, or rather the lack of it, is observed for both FR models on both datasets by looking at Figure\, \ref{fig:fdr_database} (a) and (b), as well as the FDR and FDR AUC values in Table \ref{tab:performance_ethnicity}. The FDR score varies widely over a wide range of $\tau$ in the RFW dataset. For BFW dataset, the FDR values are generally slightly higher for the smaller ArcFace r50 model, as seen in Figure\, \ref{fig:fdr_database} (b). 
However, the FDR AUC specifies the smaller model ArcFace r50 as having slightly higher fairness towards ethnicity groups compared to ArcFace r100 in both datasets. 

In Table \ref{tab:performance_gender}, we see that males perform slightly better compared to females in terms of FNMR and FMR across multiple thresholds, indicating some form of inherent gender bias in both FR models. 
Now, looking at Figure\,\ref{fig:fdr_database}(c), the FDR curve shows the slightly higher gender fairness of the larger ArcFace r100 model. 

In summary, both FR models have consistent performance trends and less-than-perfect fairness in both the gender and ethnicity demographic groups. Thus, explaining the differences in the FR model's reaction to these groups, and the consistency of this explainability, is highly relevant to understanding their behaviour.

\begin{table}
\centering
\resizebox{0.8\linewidth}{!}{%
\begin{tabular}{l|lll|lll}
\cline{2-7}
        & \multicolumn{3}{c|}{ArcFace r100}                                  & \multicolumn{3}{c}{ArcFace r50}                                    \\ \hline
       $\tau@FMR_10^x$ & \multicolumn{1}{l|}{$10^{-2}$}      & \multicolumn{1}{l|}{$10^{-3}$}      & $10^{-4}$      & \multicolumn{1}{l|}{$10^{-2}$}      & \multicolumn{1}{l|}{$10^{-3}$}      & $10^{-4}$      \\ \hline
       \hline
Gender  & \multicolumn{3}{c|}{FMR}                                           & \multicolumn{3}{c}{FMR}                                            \\ \hline
Males   & \multicolumn{1}{l|}{1.1E-2} & \multicolumn{1}{l|}{1.1E-3} & 1.2E-4 & \multicolumn{1}{l|}{1.1E-2} & \multicolumn{1}{l|}{1.1E-3} & 1.1E-4 \\ \hline
Females & \multicolumn{1}{l|}{1.6E-2} & \multicolumn{1}{l|}{1.8E-3} & 1.7E-4 & \multicolumn{1}{l|}{1.6E-2} & \multicolumn{1}{l|}{1.8E-3} & 1.8E-4 \\ \hline
\hline
        & \multicolumn{3}{c|}{FNMR}                                          & \multicolumn{3}{c}{FNMR}                                           \\ \hline
Males   & \multicolumn{1}{l|}{0.085}  & \multicolumn{1}{l|}{0.106}  & 0.131  & \multicolumn{1}{l|}{0.089}  & \multicolumn{1}{l|}{0.118}  & 0.147  \\ \hline
Females & \multicolumn{1}{l|}{0.083}  & \multicolumn{1}{l|}{0.110}  & 0.151  & \multicolumn{1}{l|}{0.091}  & \multicolumn{1}{l|}{0.128}  & 0.173  \\ \hline
\hline
FDR     & \multicolumn{1}{l|}{0.997}  & \multicolumn{1}{l|}{0.998}  & 0.990  & \multicolumn{1}{l|}{0.996}  & \multicolumn{1}{l|}{0.994}  & 0.987  \\ \hline
FDR AUC & \multicolumn{3}{l|}{0.992}                                         & \multicolumn{3}{l}{0.990}                                          \\ \hline
\end{tabular}
}
\caption{FNMR($\tau$), FMR($\tau$), and FDR($\tau$) are given per gender as subgroup, where the operational points are defined as $\tau$ at $FMR_x$. It is to note that $\tau$ is set using the entire test dataset, due to missing development set.}
\label{tab:performance_gender}
\end{table}

\subsection{Explainability of ethnic differences in FR}

In Figures \ref{fig:ArcFace_BFW_scoreCAM}, \ref{fig:ArcFace_RFW_scoreCAM}, \ref{fig:ResNet50_BFW_scoreCAM} and \ref{fig:ResNet50_RFW_scoreCAM}, the MAM of different ethnicity groups are visually very similar. This goes as well to the histogram of the MAM values comparison between the ethnic groups. This inability to see differences in the FR model's reaction to different groups, which is expected due to the demonstrated lack of fairness, is the main motivation behind our explainability tools. Rather than searching for differences in the model activation maps, we look for differences in the variation of these activations, thus the AM-V and its derivative, the D-AM-V.

The D-AM-V reveals better the spatially related difference between these demographic groups. Local areas with a higher difference in the activation variation indicate a higher difference between the way FR models see different ethnic groups. As indicated by the pipeline in Figure\,\ref{fig:pipeline}, we build our investigations always to the Caucasian group as reference. Figure\,\ref{fig:ArcFace_BFW_scoreCAM},  \ref{fig:ArcFace_RFW_scoreCAM}, \ref{fig:ResNet50_BFW_scoreCAM} and \ref{fig:ResNet50_RFW_scoreCAM} show the same shape of the D-AM-V maps for all demographics sets (E-C, A-C, and I-C) over two different datasets within the same FR model. D-AM-V maps demonstrate for Indian strong differences on the nose and outer eye corners, while for Africans the focus lies around the mouth, chin, and forehead, and for Asian on the cheeks and forehead area.  
Mapping the D-AM-V along the x- and y- direction show a higher difference in the activation variation between Caucasians (red) than non-Caucasians (blue) ethnic groups. Looking at the $s_y$ maps (in E-C, A-C, and I-C) for the forehead, nose, and chin areas, the same findings as before can be obtained while observing large gaps in the cheek areas for the Asian group, between the mouth-chin area in Africans, and between the nose region in Indians. In general, Asians show higher D-AM-V values, which is probably related to them scoring some of the worse verification performances across ethnicity groups on both FR models (see Table \ref{tab:performance_ethnicity}).

These observations are rather consistent to a large degree on both FR models and both databases (see Figures \ref{fig:ArcFace_BFW_scoreCAM}, \ref{fig:ArcFace_RFW_scoreCAM}, \ref{fig:ResNet50_BFW_scoreCAM} and \ref{fig:ResNet50_RFW_scoreCAM}). However, comparing the AM-V distributions across non-Caucasians and Caucasians for BFW in Figure\,\ref{fig:ArcFace_BFW_scoreCAM} and Figure\,\ref{fig:ResNet50_BFW_scoreCAM}, one sees a stronger variation in ArcFace r50 than in ArcFace r100, suggesting that a smaller backbone causes stronger variations in activation, which can be related to its general lower verification performance. The same trend is also observed for RFW across ArcFace r100 and ArcFace r50 in Figure\,\ref{fig:ArcFace_RFW_scoreCAM} and Figure\,\ref{fig:ResNet50_RFW_scoreCAM}. 

In summary, the observations made on the most important areas of difference in the activation variation between the different ethnic groups and the Caucasian group are to a large degree consistent for both FR models and datasets. Different demographics show characteristic patterns in terms of the D-AM-V map in comparison to Caucasians highlighting specific areas of interest across demographic groups.
In a study on the facial anthropometric differences among ethnicity \cite{10.1093/annhyg/meq007}, the authors pointed out that the chin arc and sub-nasal arc (mouth region) were on the top in terms of average change between African Americans and Caucasians, which is consistent with our observations through the eyes of FR models and confirms the validity of our explainability tools. Unfortunately, the study did not include information about the Indian and Asian groups.

\subsection{Explainability of gender differences in FR }


Figure\,\ref{fig:gender_BFW_ScoreCAM} shows the group characteristics of the AM maps for the gender aspect. Only considering the shape of the MAM, no clear deviation is visible between both genders. However, when we focus on higher-order analysis, such as the D-AM-V, we observe a differentiation in the forehead and chin regions. 
This is rather consistent for both FR models.
Very interestingly, in a study that analyse the facial areas that mostly affect the human judgment on the gender of the face \cite{doi:10.1068/p220131}, the chin and brow (forehead) areas came on top. Which, to some degree points out the sanity of our proposed explainability tools. 
Additional study on the facial anthropometric differences among genders \cite{10.1093/annhyg/meq007} pointed out that the chin arc and frontal arc (forehead) were on the top in terms of average change between females and males' facial measurements.


One limitation of our study is the number of experimental variations that can fit in such a work. We chose to build variations in the FR model architecture and the dataset to avoid "bias" outcomes in these regards. However, interesting questions regarding the explainability tool's outcome across FR training losses \cite{DBLP:conf/cvpr/DengGXZ19,DBLP:journals/corr/abs-2109-09416,DBLP:journals/pr/BoutrosDKK22}, network architectures \cite{DBLP:conf/cvpr/HeZRS16,DBLP:conf/icb/BoutrosDFKK21,DBLP:conf/iccvw/YanZXZWS19,DBLP:journals/access/BoutrosSKDKK22}, FR training datasets \cite{guo2016ms,DBLP:conf/fgr/CaoSXPZ18}, the set of analysed demographics (or even non-demographic) variations \cite{9534882}, the combined analyses of demographic groups (e.g. African females in comparison to Indian males), and the pairing of the compared demographic groups (we chose the top performer as a reference here), are yet to be explored.

\section{Conclusion} 

In this work, we aimed at explaining the difference in the perspective of FR models between different demographic groups. 
Towards that, we presented a set of explainability tools visualizing the ethnic and gender differences for the underlying FR models. In general, both considered FR models show ethnic bias in both datasets in terms of unequal verification performance in different demographic groups. 
Our tools and analyzing the results on two datasets and two FR models pointed out certain regions that might cause the FR model's behavior differences between certain ethnic groups and the Caucasian ethnicity on one hand, and between males and females on the other hand.
Interestingly, the outcome is, to a large degree, consistent with the available clues from facial anthropometric differences studies and studies on the human judgment of gender from faces.

\textbf{Acknowledgements:}
This research work has been funded by the German Federal Ministry of Education and Research and the Hessian Ministry of Higher Education, Research, Science and the Arts within their joint support of the National Research Center for Applied Cybersecurity ATHENE.
{\small
\bibliographystyle{ieee}
\bibliography{egbib}
}

\end{document}